# ParseTalk ABOUT TEXTUAL ELLIPSIS

**Michael Strube** and **Udo Hahn**

 Computational Linguistics Research Group
Freiburg University
D-79085 Freiburg, Germany
email: {strube,hahn}@coling.uni-freiburg.de

**Abstract**
A hybrid methodology for the resolution of text-level ellipsis is presented in this paper. It incorporates conceptual proximity criteria applied to ontologically well-engineered domain knowledge bases and an approach to centering based on functional topic/comment patterns. We state text grammatical predicates for ellipsis and then turn to the procedural aspects of their evaluation within the framework of an actor-based implementation of a lexically distributed parser.

## 1 Introduction

The work reported in this paper is part of a large-scale text understanding system for knowledge acquisition from German expository texts. Text phenomena are a particularly challenging issue for the design of appropriate parsers, since lacking recognition facilities either result in referentially incohesive or, even worse, invalid text knowledge representations. In a previous paper (Strube and Hahn, 1995), we have already dealt with text-level anaphora (e.g., "Jack owns *a car*. *It* cost him \$35,000."), the resolution of which contributes to the construction of (referentially) valid text knowledge bases. In this paper we propose a methodology for the resolution of text-level ellipsis yielding (referentially) cohesive text knowledge bases. The phenomena we address (also called functional anaphora) can be illustrated by the following sentence pair: "Jack owns a *car*. *The tires [of the car]* need to be changed." ("[...]" indicates material deleted from the surface expression). The approach to text ellipsis resolution we propose integrates language-independent conceptual (distance measure) and language-dependent functional (topic/comment) constraints based on the centering approach (Grosz et al., 1995).

We explicitly exclude two terminologically related problems from our study. First, we restrict the consideration of ellipses to their textual form, i.e., one that extends over formal sentence boundaries. This excludes, in particular, any constructions which build on coordination and corresponding elision phenomena within the sentence (e.g., "*Jack owns a car, [and Jack owns] a house, and [Jack owns] a record shop.*"). We also exclude cases where several lexical "traces" signal elliptical expressions, e.g., phenomena underlying VP ellipsis (e.g., "Jack *owns a car*, and *so does* John *[own a car]*."), an issue of particular relevance for the English language but almost irrelevant for others such as German. These forms of ellipses are usually explained in terms of structural, i.e., syntactic phenomena, viz. the application of proper deletion, recoverage or copying rules for "parallel" constructions (cf., e.g., Hardt (1992a; 1992b), Kehler (1993), Grover et al. (1994)).

Second, ellipses in written texts must clearly be distinguished from elliptical constructions as they occur in question-answering dialogues (e.g., Q: "What *is your hobby*?" A: "Playing jazz music *[is my hobby]*."; for a treatment of that issue, cf., e.g., Weischedel and Sondheimer (1982) and Carbonell (1983), who emphasize syntactic and semantic strategies for ellipsis resolution, while Carberry (1989) elaborates on multiple pragmatic criteria involving beliefs, plans and goals). While QA ellipsis, when isolated from its discourse setting, often tends to be ungrammatical or at least fragmentary at the surface level, textual ellipsis is characterized by entirely grammatical sentences which only lack explicit reference to the discourse entities already available from the context.

Surprisingly little efforts have already been spent on the design of computational models for the analysis of textual ellipsis (cf. Section 7), although these constructions occur at significant rates in expository texts. Fraurud (1990) carried out an experimental study on the distribution of (in)definite NPs in such texts and found out that 36% of their occurrences can be classified as cases of ellipsis, while another 36% belong to the category of anaphora.

## 2  Constraints on Textual Ellipsis

Textual ellipsis – at the conceptual level – relates an elliptical expression to its antecedent by conceptual attributes (or roles) associated with that antecedent (see (1) below). It thus complements the phenomenon of nominal anaphora, which are related to their antecedent in terms of conceptual generalization (cf. Strube and Hahn (1995)).

(1) Compaq bestückt den LTE-Lite/25 mit einem PCI-Motherboard. *Die CPU* hat eine Taktfrequenz von 50 Mhz.
*(Compaq equips the LTE-Lite/25 with a PCI-motherboard. The CPU comes with a clock frequency of 50 Mhz.)*

We call this phenomenon textual ellipsis because in the second sentence a conceptual entity that relates the topic of this sentence to discourse elements mentioned in the preceding one is missing. Hence, the appropriate conceptual link must be inferred in order to establish the cohesion of the whole discourse (for an early statement of that idea, cf. Clark (1975)). In (1) the information is missing that the *CPU* is part of the *(PCI-)motherboard*. This apparent relation can only be inferred if conceptual knowledge about the domain is available. It is obvious (see Fig. 1)[1] that the concept CPU is bound in a direct *part-of*-type relation, viz. *has-cpu*, to the concept MOTHERBOARD, while its partonomic relation to the instance LTE-LITE-25 is not so tight; a conceptual relationship between CPU and COMPAQ is excluded at the conceptual level, since they are not linked via any *part-of*-type conceptual role.

Nevertheless, *part-of*-type conceptual roles are far too unconstrained to properly discriminate among several possible antecedents in the preceding discourse context. We therefore propose a basic heuristic for conceptual proximity which takes the path length between concept pairs into account. It is based on the common distinction between concepts and relations/roles in classification-based terminological reasoning systems (cf. MacGregor (1991) for a survey). Conceptual proximity takes only conceptual roles into consideration, while it does not consider the generalization hierarchy between concepts. The heuristic can be phrased like the following: If connecting role paths between the concepts denoted by possible antecedents and an elliptical expression exist via one or more conceptual relations (roles), that particular role composition is preferred for ellipsis resolution whose path contains the least number of roles. If several connected role paths of equal length exist, then functional constraints which are based on topic/comment patterns apply for the selection of the proper antecedent. Only at this level grammatical information from the preceding sentence is brought into play (for a more precise statement in the terms of the underlying grammar, cf. Table 5 in Section 5).

---

[1] The following notational conventions apply to the knowledge base for the information technology domain (see Fig. 1): Angular boxes from which double arrows emanate contain instances (e.g., LTE-LITE-25), while rounded boxes contain generic concepts (e.g., NOTEBOOK). Directed unlabelled links relate concepts via the *isa* relation (e.g., NOTEBOOK and COMPUTER-SYSTEM), while links labelled with an encircled square represent conceptual roles (in Fig. 1 only definitional roles occur) whose names and value constraint expressions are attached to each circle (e.g., COMPUTER-SYSTEM - *has-central-unit* - CENTRAL-UNIT, with small italics emphasizing the role name). Note in particular that any subconcept or instance inherits the conceptual attributes from its superconcept or concept class (this is not explicitly shown in Fig. 1).

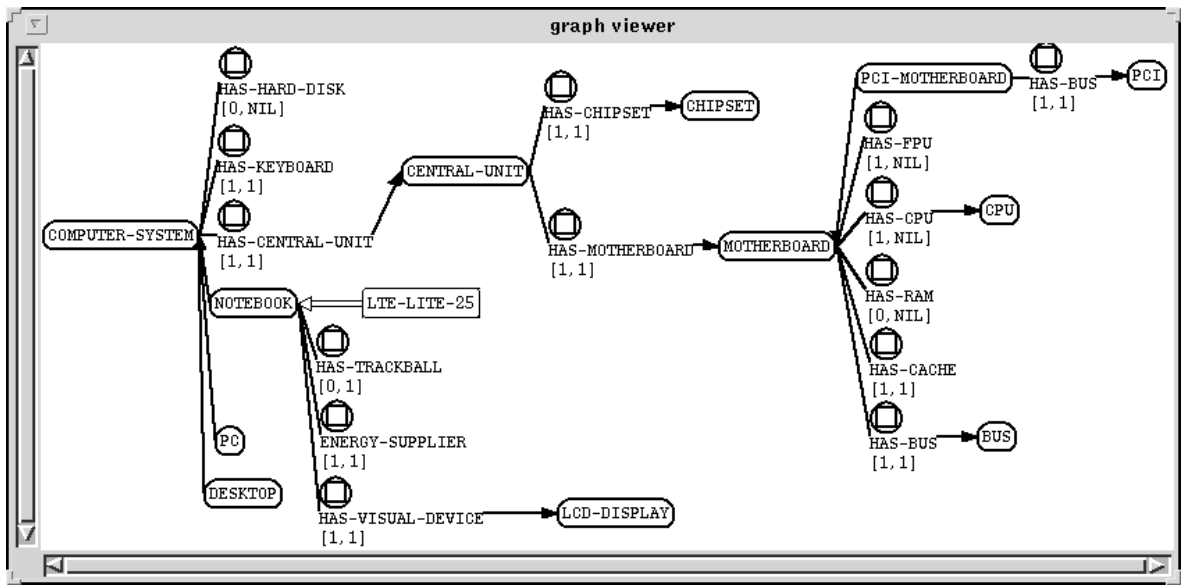

Figure 1: Fragment of the Underlying Domain Knowledge Base

Given sentence (1) and Fig. 1, according to the convention above PCI-MOTHERBOARD is conceptually most proximate to the elliptical occurrence of CPU (due to the direct conceptual role between MOTHERBOARD – *has-cpu* – CPU with unit length 1), while the relationship between LTE-LITE-25 and CPU exhibits a greater conceptual distance (counting with unit length 3, due to the triple composition of roles between COMPUTER-SYSTEM – *has-central-unit* – CENTRAL-UNIT – *has-motherboard* – MOTHERBOARD – *has-cpu* – CPU).

## 3 Epistemological Engineering for Ellipsis Resolution

Metrical criteria incorporating path connectivity patterns in network-based knowledge bases (i.e., concept graphs) have often been criticized for lacking generality and introducing *ad hoc* criteria likely to be invalidated when applied to different domain knowledge bases (DKB). The crucial point about the presumed unreliability of path-length criteria seems to address the apparent problem how the topology of such a network system can be "normalized" such that formal distance measures relate to intuitively plausible conceptual proximity judgments. Though we have no formal solution for this correspondence problem, our proposal tries to eliminate structural idiosyncrasies by postulating two *knowledge engineering (KE) principles*:

1. **Clustering into Basic Categories.** The specification of the upper level of the ontology of some domain (e.g., information technology (IT)) should be based on a stable set of abstract, yet domain-oriented *ontological categories* inducing an *almost complete partition* on the entities of the domain at a *comparable level of generality* (e.g., hardware, software, companies in the IT world). Each specification of such a basic category and its taxonomic descendents constitutes the common ground for what Hayes (1985) calls *clusters* and Guha and Lenat (1990) refer to as *micro theories*, i.e., self-contained descriptions of conceptually related proposition sets about a reasonable portion of the commonsense world *within* a single knowledge base partition (context, subtheory).

2. **Balanced Deepening.** The specification of the lower levels of that ontology dealing with concrete objects of the domain (e.g., notebooks, laser printers, hard disks in the IT world) should be carefully balanced, i.e., the extraction of attributes for any particular category should proceed at a uniform degree of detail at each decomposition level. This requirement should guarantee that any subworld has the *same level of representational granularity*, although the representational granularities might differ among various subworlds (associated with different basic categories).

These principles are but a bottom-line for a discipline that might be called *epistemological engineering*. It is currently emerging from several attempts to develop a formal methodology for knowledge acquisition (cf. Alexander et al. (1988)), but still relies on the provision of experiential guidelines for building concrete, non-toy DKBs in a terminological (cf. Brachman et al. (1991, Section 14.5), Gates et al. (1989, Section 1) or Monarch and Nirenburg (1990)) or predicate calculus language framework (Hobbs, 1984). The problems one encounters in that area are rooted in fundamental philosophical problems and are rearticulated by all those involved in a metatheory of knowledge representation in the artificial intelligence camp. McCarthy (1977), for example, stresses questions of observational availability of data and correspondence relations between observable data and their proper formal representation.

These are not just abstract philosophical debates, as the problems under consideration turn up in every KE enterprise. Davis et al. (1993, pp. 19-21) convincingly argue that we are caught in a plethora of *ontological commitments* which accumulate, at least, at three layers — the choice of a particular representation format (logic, nets, frames, etc.), the force of the major representation constructs of this framework (e.g., prototypicality, defaults, and taxonomic hierarchies as far as frame representations are concerned), and, finally, the knowledge engineering level (which we specifically address by the KE principles stated above). At this level, human conceptual bias results from (unconscious) selectivity of observation. But also the perspective one chooses to solve the representation problem by (consciously) focusing on the "relevant" issues adds further bias: Which knowledge items should be included in the actual representation structures? Where do they appear in the hierarchy (again, if object-centered representations are chosen)?, etc.

The ellipsis resolution problem, nevertheless, incorporates any of these commitment layers and even projects solutions worked out at the *knowledge layer* on the data structure or *symbol layer* of representations. By this, we mean the abstract implementation of knowledge representation structures in terms of graphs and their path connectivity patterns. At this level, however, we have reasons to assume that the proximity metric on which we build makes sense. We here draw on early work from cognitive psychologists such as Quillian (1969) and Rips et al. (1973), or more recent research in the parsing domain proper, e.g., by Charniak (1986). Their experiments provide ample evidence that the definition of proximity in semantic networks in terms of the traversal of typed edges (e.g., only via generalization or via attribute links) and the corresponding counting of nodes that are passed on that traversal is not only common practice but also methodologically valid for computing conceptual distances.

The KE principles mentioned above are supplemented by the following linguistic regularities which hold for textual ellipsis:

1. **Adherence to a Focused Context.** Valid antecedents of elliptical expressions occur *within subworld boundaries* (in technical terms, they remain within a single knowledge base context, micro theory, etc.). Given the above KE constraints (in particular the one requiring each subworld to be characterized by the same degree of conceptual density), path length criteria make sense for estimating the conceptual proximity between concepts. Moreover, only links of a certain type are considered for traversals, viz. those covered by the *part-of* relation.

2. **Limited Path Length Inference.** Valid pairs of possible antecedents and elliptical expressions denote concepts in the DKB whose conceptual relations (role chains) are constructed on the basis of rather *restricted path length* units (in our experiments, no valid chain ever exceeded unit length 5). This corresponds to the implicit requirement that these role chains must be efficiently computable.

Remaining within a focal subworld, constraining the relation types for graph traver-sal and limiting the number of nodes being passed — these are the major constraints for the creation (and interpretation) of such elliptical relations by the text producer (understander). They must be met to preserve textual cohesion.

| anaphora $>_{TC_{base}}$ head rheme $>_{TC_{base}}$ right edge rheme $>_{TC_{base}}$ head theme $>_{TC_{base}}$ head unmarked |
| --- |
| pronominal anaphor $>_{TC_{anatype}}$ possessive pronoun $>_{TC_{anatype}}$ nominal anaphor $>_{TC_{anatype}}$ textual ellipsis |
| subject $>_{TC_{anafunc}}$ direct object $>_{TC_{anafunc}}$ indirect object $>_{TC_{anafunc}}$ adjunct |

Table 1: Functional Ranking on the $C_f$ for German according to Topic/Comment Patterns

## 4 Centering Principles for German

Conceptual criteria are of tremendous importance, but they are not sufficient for proper ellipsis resolution. Additional criteria have to be supplied in the case of equal role length for several alternative antecedents. We therefore incorporate into our model various functional criteria in terms of topic/comment patterns which originate from the (dependency) structure of the preceding sentence. The organizational framework for this type of information is provided by the well-known *centering* mechanism (Grosz et al., 1995). Accordingly, we distinguish each utterance's backward-looking center ($C_b(U_n)$) and its forward-looking-centers ($C_f(U_n)$). The ranking imposed on the elements of the $C_f$ reflects the assumption that the first element of $C_f(U_n)$ is the most preferred antecedent of an anaphoric (or elliptical) expression in $U_{n+1}$, while the remaining elements are ordered according to decreasing preference for establishing referential links.

The main difference between Grosz et al.'s work and our proposal concerns the criteria for ranking the forward-looking centers. While Grosz et al. (1995) assume (for the English language) that *grammatical* roles are a major determinant for the ranking on the $C_f$, we claim that the major determinant for German – a language with relatively free word order – is the *functional* information structure of the sentence in terms of topic/comment patterns. Accordingly, the *topic* denotes the given information which occurs at the beginning of the sentence, while the *comment* denotes the new information which is given at the end of the sentence. All phrases intervening topic and comment are called *unmarked*, because they are irrelevant for the information structure of a sentence (cf. Fig. 2 for an abstract configuration schema in terms of dependency grammar). Note that there are some exceptions to this general statement, which relate to syntactic phenomena like coordination, copula sentences, interrogative clauses, some types of subordinate clauses, etc.

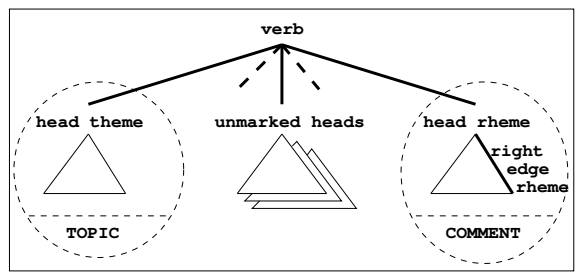

Figure 2: Abstract Configuration Schema for Topic/Comment Patterns

Not only are topic/comment patterns relevant for the ranking on the $C_f$ but also is it important whether an element of the sentence is anaphoric or not. Anaphoric elements are generally ranked higher than any non-anaphoric elements, irrespective of the topic/comment structure of the sentence in which they occur (cf. Hajičová et al. (1992)).

The rules holding for the ranking on the $C_f$ for German are summarized in Table 1. They are organized at three layers. At the top level, $>_{TC_{base}}$ denotes the basic relation for the overall ranking of topic/comment (TC) patterns. Accordingly, any anaphoric expression in the preceding sentence $U_n$ is given the highest preference as a potential antecedent of an anaphoric (or elliptical) expression in $U_{n+1}$; the other types of functional configurations, viz. head rheme, right edge rheme, head theme, or unmarked head(s), are accessible in the given decreasing preference order. Our studies on expository texts in Ger-man have shown that there are no anaphora whose antecedents occur as modifiers except of those at the right edge of the rheme; therefore they are included in $C_f$ (cf. Fig. 2).

The middle and the bottom-level of Table 1 are constituted by $>_{TC_{anatype}}$ and $>_{TC_{anafunc}}$ which denote preference relations dealing exclusively with multiple occurrences of (resolved) anaphora in the preceding sentence. $>_{TC_{anatype}}$ distinguishes among *different forms* of anaphora (viz., pronominal, possessive, nominal and ellipsis form) and their associated preference order, while $>_{TC_{anafunc}}$ covers the functionally based preference order for multiple occurrences of the *same type* of anaphora (i.e., whether they occur as subject, direct object, indirect object or adjunct).

Given these basic relations, we may now formulate the composite relation $>_{TC}$ (Table 2). It states the conditions for the comprehensive ordering of items on $C_f$ ($x$ and $y$ denote lexical heads).

```
>_{TC} := { (x, y) |
  if x and y both represent the same type
     of anaphora
  then the relation >_{TC_{anafunc}} applies to x and y
  else if x and y both represent different forms
     of anaphora
  then the relation >_{TC_{anatype}} applies to x and y
  else the relation >_{TC_{base}} applies to x and y }
```

Table 2: Global Topic/Comment Relation

## 5 Grammatical Predicates for Textual Ellipsis

We here build on the *ParseTalk* model of dependency grammar, a fully lexicalized grammar theory which employs default inheritance for lexical hierarchies (Bröker et al., 1994; Hahn et al., 1994)). The grammar formalism is based on dependency relations between lexical heads and modifiers at the sentence level. The dependency specifications allow a tight integration (not a mixture!) of linguistic knowledge (grammar) and conceptual knowledge (domain model), thus making powerful terminological reasoning facilities directly available for the parsing process (cf. also Hajičová (1987) in support of this view). Accordingly, syntactic analysis and semantic interpretation are closely coupled.

This exposition of the *ParseTalk* grammar framework is tailored to the requirements of the resolution of textual ellipses. We assume the following conventions to hold: $\mathcal{C}$ = {Word, Nominal, Noun, DetDefinite,...} denotes the set of word classes, and $isa_\mathcal{C}$ = {(Nominal, Word), (Noun, Nominal), (DetDefinite, Nominal),...} $\subset \mathcal{C} \times \mathcal{C}$ denotes the subclass relation which yields a hierarchical ordering among these classes. The concept hierarchy consists of a set of concept names $\mathcal{F}$ = {COMPUTER-SYSTEM, NOTEBOOK, CENTRAL-UNIT,...} (cf. Fig. 1) and a subclass relation $isa_\mathcal{F}$ = {(NOTEBOOK, COMPUTER-SYSTEM), (PCI-MOTHERBOARD, MOTHERBOARD),...} $\subset \mathcal{F} \times \mathcal{F}$. The set of role names $\mathcal{R}$ = {*has-part, has-cpu,...*} contains the labels of admitted conceptual roles. The relation *permit* $\subset \mathcal{F} \times \mathcal{R} \times \mathcal{F}$ characterizes the range of possible conceptual roles among concepts, e.g., (MOTHERBOARD, *has-cpu*, CPU) $\in$ *permit*. Furthermore, object.attribute denotes the value of the property attribute at object, while head denotes a structural relation within dependency trees, viz. $x$ being the head of $y$.

The resolution of textual ellipsis is based on two major criteria, a structural and a conceptual one. The structural condition is embodied in the predicate *isPotentialEllipticAntecedent* (Table 3). An elliptical relation between two lexical items is here restricted to pairs of nouns. The elliptical phrase which occurs in the $n$-th sentence is restricted to a definite NP and the antecedent must be one of the forward-looking centers of the preceding sentence.

```
isPotentialEllipticAntecedent (x, y, n) :⇔
  x isa_C* Noun
  ∧ y isa_C* Noun
  ∧ ∃ z: (y head z ∧ z isa_C* DetDefinite)
  ∧ y ∈ U_n
  ∧ x ∈ C_f(U_{n-1})
```

Table 3: Determining a Potential Elliptical Antecedent

The function *ProximityScore* (Table 4) captures the basic conceptual condition in terms of the distance between two concepts.

More specifically, there must be a connected path linking the two concepts under consideration via a chain of conceptual roles.

$$\text{ProximityScore (from-concept, to-concept)}$$
$$:= \begin{cases} n \in I\!\!N \\ \quad \text{if } \exists\ x_0, ..., x_n \in \mathcal{F}\colon \exists\ r_0, ..., r_{n-1} \in \mathcal{R}\colon \\ \quad\quad x_0 = \text{from-concept} \land x_n = \text{to-concept} \\ \quad\quad \land\ \forall\ i \in [0, n\text{-}1]\colon (x_i, r_i, x_{i+1}) \in permit \\ \infty \text{ else} \end{cases}$$

Table 4: Determining the Conceptual Distance between Two Concepts

Finally, the predicate *PreferredConceptualBridge* (Table 5) combines both criteria. A lexical item $x$ is determined as the proper antecedent of the elliptical expression $y$ if it is a potential antecedent and if there exists no alternative antecedent $z$ whose *ProximityScore* either is below that of $x$ or, if their *ProximityScore* is identical, whose strength of preference under the TC relation is higher than that of $x$:

$$\text{PreferredConceptualBridge (x, y, n)} :\Leftrightarrow$$
$$\text{isPotentialEllipticAntecedent (x, y, n)}$$
$$\land\ \neg\exists\ z : \text{isPotentialEllipticAntecedent (z, y, n)}$$
$$\land\ (\ \text{ProximityScore (z.concept, y.concept)}$$
$$\quad < \text{ProximityScore (x.concept, y.concept)}$$
$$\lor\ (\ \text{ProximityScore(z.concept, y.concept)}$$
$$\quad = \text{ProximityScore(x.concept, y.concept)}$$
$$\land\ z >_{TC} x\ )\ )$$

Table 5: Determining the Preferred Conceptual Bridge for an Elliptical Expression

The mechanism we provide for the resolution of text-level ellipses is strongly rooted in the structural properties of KL-ONE-type terminological knowledge representation languages (MacGregor, 1991). Its focus is on aggregation or mereological relations (the most general form being *part-of*, but more refined conceptual roles usually must be supplied)[2]. Our metrical criterion favors the most proximate non-generalization-based link between the concept denoted by some already available discourse entity (the antecedent) and the concept denoted by the currently considered lexical item (the elliptical item). In case the distances are of equal length functional considerations complement the conceptual ones on the basis of the information structure of the preceding sentence.

## 6 Text Cohesion Parsing: Ellipsis Resolution

The actor computation model (Agha and Hewitt, 1987) provides the background for the procedural interpretation of lexicalized grammar specifications, as those given in the previous section, in terms of so-called word actors (Schacht et al., 1994). *Word actors* combine object-oriented features with concurrency and thus provide a methodologically clean platform for inherently parallel, lexically distributed computations. The model assumes word actors to communicate via asynchronous message passing. An actor can only send messages to other actors it knows about, its so-called acquaintances.

The resolution of textual ellipsis depends on the results of the resolution of nominal anaphora as well as on the termination of the semantic interpretation of the sentence. It will only be triggered at the occurrence of phrase $P$

- when $P$ is non-anaphoric, and
- when $P$ is not connected at the conceptual level (via a *has-part*-type relation) to some referent denoted in the current sentence.

The protocol level of text analysis encompasses the procedural interpretation of the grammatical constraints from Section 5. A *SearchTextEllipsisAntecedent* message is triggered if a *SearchNomAntecedent* message (intended to account for the resolution of nominal anaphora) quits without success and the semantic interpretation did not produce a proper (partonomic) relation at the conceptual level of representation for the elliptical phrase. The protocol for establishing cohesive links based on the recognition of *textual ellipsis* consists of two phases:

---

[2] The need for complex semantic data structures for proper ellipsis resolution has also been recognized by other computational linguists, e.g., by Kehler (1993), whose discourse copying algorithm uses a Davidsonian-style event representation which is close to the notion of frames. However, his use of event structures, or similarly, the event types of the unification discourse grammar framework to which Grover et al. (1994) refer are rather paraphrases of common case role labels than sophisticated conceptual attributes (both papers aim at VP ellipsis!), and thus lack the level of conceptual differentiation needed to adequately cope with textual ellipsis.

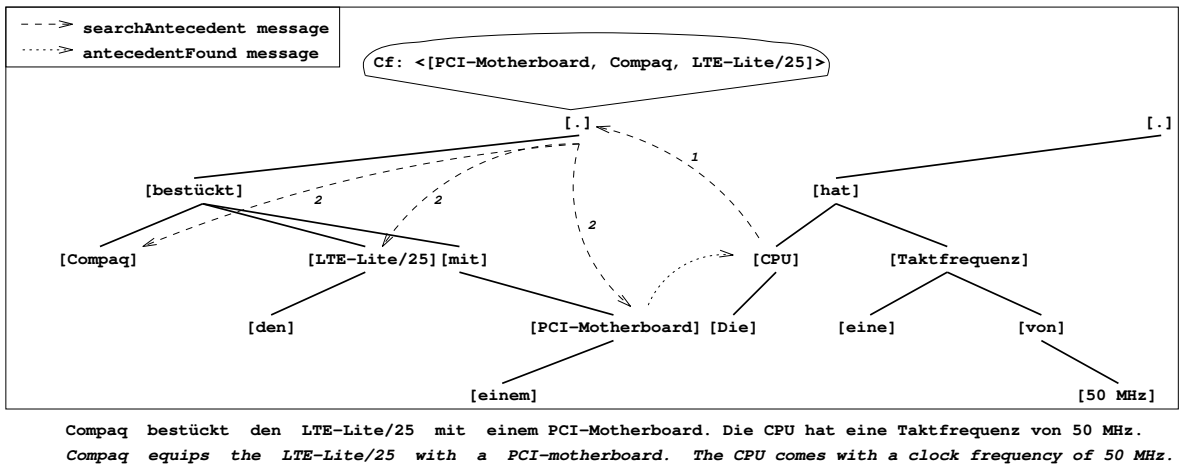

Figure 3: Sample Parse for Text Ellipsis Resolution

1. In *phase 1*, the message is forwarded from its initiator to the sentence delimiter of the *preceding sentence*, where its state is set to *phase 2*.

2. In *phase 2* the sentence delimiter's acquaintance $C_f$ is tested for the predicate *PreferredConceptualBridge*.

Note that only nouns and pronouns are capable of responding to the *SearchTextEllipsisAntecedent* message and of being tested as to whether they fulfill the required criteria for an elliptical relation. If the ellipsis predicate *PreferredConceptualBridge* succeeds, the determined antecedent sends a *TextEllipsisAntecedentFound* message to the initiator of the *SearchTextEllipsisAntecedent* message. Upon receipt of the *AntecedentFound* message, the discourse referent of the elliptical expression is conceptually related to the antecedent's concept via a *has-part*-type relation, thus preserving cohesion at the conceptual level of text propositions. We will now illustrate (cf. Fig. 3) the protocol for establishing elliptical relations, referring to the already introduced text fragment (1) which is repeated at the bottom line of Fig. 3.

The second sentence of (1) contains the definite noun phrase *die CPU*. Since *CPU* does not subsume any word at the conceptual level in the preceding text (cf. Fig. 1), the anaphora test fails; the definite noun phrase *die CPU* has also not been integrated in terms of a partonomic relation into the conceptual representation of the sentence as a result of the semantic interpretation. As a consequence, a *SearchTextEllipsisAntecedent* message is created by the word actor for *CPU*. That message is sent directly to the sentence delimiter of the previous sentence, where the predicate *PreferredConceptualBridge* is evaluated for the acquaintance $C_f$. As one of the relevant heads can be tested successfully (the corresponding concept PCI-MOTHERBOARD is related to CPU via the role *has-cpu*), a *TextEllipsisAntecedentFound* message is sent to the initiator of the *SearchAntecedent* message. An appropriate update links the corresponding concepts via the role *has-cpu* and, thus, cohesion is established at the conceptual level of the text knowledge base.

In order to illustrate our approach under slightly varying conditions, let us discuss examples of ellipsis resolution with focus on the DKB fragment depicted in Fig. 1. We abstract from the event-oriented description of the parsing process itself and concentrate instead on the computations being performed within the knowledge representation system. Consider text fragment (1) again. The relevant knowledge base (KB) operations (see the listing in Table 6) are caused by the evaluation of the predicate *PreferredConceptualBridge* and are performed on three in-

```
(preferred-cb (PCI-MOTHERBOARD-0004
COMPAQ LTE-LITE-25) CPU-0005)

PCI-MOTHERBOARD-0004 - depth:1 TRUE

==> (PCI-MOTHERBOARD-0004 HAS-CPU CPU-0005)
```

Table 6: Transcript from the Domain Knowledge Base for Text Fragment (1)

stances, viz. PCI-MOTHERBOARD, COMPAQ, and LTE-LITE-25, which form the current forward-looking centers. The terminological reasoner attempts to determine a role chain between one of these instances and CPU. Only PCI-MOTHERBOARD passes this test successfully. Both concepts can be linked via the *has-cpu* role at unit length (depth) 1.

Consider a slight variation of text fragment (1): If *PCI-Motherboard* in (1) is replaced by *LCD-Display* (as in (2)) the result of the ellipsis resolution differs from the previous example. Since due to conceptual constraints LCD-DISPLAY cannot be considered a proper antecedent of CPU, LTE-LITE-25 is selected as the only valid antecedent. The corresponding conceptual link (see the KB listing in Table 7) consists of the relation composition LTE-LITE-25 – *has-central-unit* – CENTRAL-UNIT – *has-motherboard* – MOTHERBOARD – *has-cpu* – CPU (having depth 3).

(2) Compaq bestückt den LTE-Lite/25 mit einem LCD-Display. *Die CPU hat eine Taktfrequenz von 50 Mhz.*
(*Compaq equips the LTE-Lite/25 with a LCD-display. The CPU comes with a clock frequency of 50Mhz.*)

```
(preferred-cb (LCD-DISPLAY-0008 COMPAQ
LTE-LITE-25) CPU-0009)

LCD-DISPLAY-0008 - depth:1 NIL
COMPAQ - depth:1 NIL
LTE-LITE-25 - depth:1 NIL
LCD-DISPLAY-0008 - depth:2 NIL
COMPAQ - depth:2 NIL
LTE-LITE-25 - depth:2 NIL
LCD-DISPLAY-0008 - depth:3 NIL
COMPAQ - depth:3 NIL
LTE-LITE-25 - depth:3 TRUE

==> (LTE-LITE-25 (COMPOSE HAS-CENTRAL-UNIT
HAS-MOTHERBOARD HAS-CPU) CPU-0009)
```

Table 7: Transcript from the Domain Knowledge Base for Text Fragment (2)

## 7 Comparison with Related Approaches

As far as text-level processing is concerned, the framework of DRT (Kamp and Reyle, 1993), at first sight, constitutes a particular strong alternative to our approach. The complex machinery of DRT, however, might work well for anaphora, but would fail if non-anaphoric, e.g., elliptical text phenomena had to be interpreted (but see Wada (1994) for a recent attempt to deal with restricted forms of ellipsis in the DRT context). This shortcoming is simply due to the fact that DRT is basically a semantic theory, not a full-fledged model for text understanding. In particular, it lacks any systematic connection to well-developed reasoning systems accounting for conceptual knowledge and specific problem-solving models which underlie the chosen domain.

As far as proposals for the analysis of textual ellipsis are concerned, none of the standard grammar theories (e.g., HPSG, LFG, GB, CG, TAG) covers this issue. This is not surprising at all, as their advocates pay almost no attention to the text level of linguistic description (with the exception of several forms of anaphora) and also do not take conceptual criteria as part of grammatical descriptions seriously into account. Conventional grammar-based approaches seem entirely infeasible unless one were willing to duplicate the knowledge which has been gathered in a DKB at the grammar level in terms of, say, a highly diversified "case" grammar system (e.g., as advocated by Kehler (1993) and Grover et al. (1994)). Unfortunately, this leaves us with the same problems (as those already discussed in Section 3) under a different cover, since relations among different (sub)types of cases then had to be scrutinized.

Actually, only few systems exist which deal with textual ellipsis. Those which take care of it do so in an often *ad hoc* way. As an example, consider the PUNDIT system (Palmer et al. 1986) which provides a rough implementation solution for a particular domain. This work shares a lot of similarities with our approach (e.g., the use of focus mechanisms (Grosz and Sidner, 1986)). But we consider our proposal superior, since it provides a more general, implementation-independent treatment at the grammar level. The approach reported in this paper also extends our own previous work on textual ellipses (Hahn, 1989) by the incorporation of an elab-orated model of functional preferences on $C_f$ elements which constrains the set of possible antecedents according to topic/comment patterns.

## 8 Conclusion

In this paper, we have outlined a model of text ellipsis parsing. It considers conceptual criteria to be of primary importance and provides a proximity measure in order to assess various possible antecedents for consideration of proper bridges (Clark, 1975) to elliptical expressions. In addition, functional constraints based on topic/comment patterns contribute further restrictions on proper elliptical antecedents. The particular advantage of our approach lies in the integrated treatment of text-level ellipsis within a single coherent grammar format.

The anaphora resolution module (Strube and Hahn, 1995) and the ellipsis handler have both been implemented in Smalltalk as part of a comprehensive text parser for German. Besides the information technology domain, experiments with our parser have also been successfully run on medical domain texts, thus indicating that the heuristics we have been developing are not bound to a particular domain. The current lexicon contains a hierarchy of approximately 70 word class specifications with nearly 2.500 lexical entries and corresponding concept descriptions available from the LOOM knowledge representation system (MacGregor and Bates, 1987) — 650 and 400 concept/role specifications for the information technology and medicine domain, resp.

**Acknowledgments.** We would like to thank our colleagues in the $\mathcal{CLIF}$ group who read earlier versions of this paper. In particular, improvements are due to discussions we had with N. Bröker, K. Markert, S. Schacht, K. Schnattinger, and S. Staab. This work has been funded by *LGFG Baden-Württemberg* (1.1.4-7631.0; M. Strube) and a grant from *DFG* (Ha 2907/1-3; U. Hahn).